\documentclass[11pt]{article}

\usepackage[pagebackref=true]{hyperref} 
\renewcommand*{\backrefalt}[4]{%
  \ifcase #1 %
    \relax 
  \or
    (cited on page~#2)%
  \else
    (cited on pages~#2)%
  \fi%
}

\usepackage{fullpage,times,url,bm}

\usepackage{amsthm,amsfonts,amsmath,amssymb,epsfig,color,float,graphicx,verbatim}
\usepackage{algorithm,algorithmic}
\usepackage{bbm}
\usepackage[square,numbers]{natbib}

\usepackage{graphicx}
\usepackage{subcaption}

\usepackage{array}

\usepackage{hyperref}
\hypersetup{
	colorlinks   = true, 
	urlcolor     = blue, 
	linkcolor    = blue, 
	citecolor   = black 
}

\usepackage[T1]{fontenc}

\newcommand{\figref}[1]{Figure~\ref{#1}}
\renewcommand{\eqref}[1]{Eq.~(\ref{#1})}

\usepackage{amssymb}

\usepackage{bbm}

\newcommand{\stam}[1]{}

\usepackage{mathtools}

\newcommand{\be}{\mathbf{e}}
\newcommand{\bx}{\mathbf{x}}
\newcommand{\bw}{\mathbf{w}}

\newcommand{\bu}{\mathbf{u}}

\newcommand{\btheta}{{\boldsymbol{\theta}}}

\newcommand{\co}{{\cal O}}

\newcommand{\cl}{{\cal L}}

\newcommand{\cn}{{\cal N}}
\newcommand{\calr}{{\cal R}}

\DeclareMathOperator*{\argmin}{argmin}

\newcommand{\reals}{{\mathbb R}}

\newcommand{\zero}{{\mathbf{0}}}
\newcommand{\one}{{\mathbf{1}}}

\newcommand{\inner}[1]{\langle #1 \rangle}
\newcommand{\norm}[1]{\left\|#1\right\|}

\makeatletter
\newcommand{\printfnsymbol}[1]{%
  \textsuperscript{\@fnsymbol{#1}}%
}
\makeatother

\title{On the Implicit Bias in Deep-Learning Algorithms}

\author{Gal Vardi\\
TTI-Chicago and Hebrew University\\ 
\texttt{galvardi@ttic.edu}
}

\date{}

\begin{document}

\maketitle

\begin{abstract}
    Gradient-based deep-learning algorithms exhibit remarkable performance in practice, but it is not well-understood why they are able to generalize despite having more parameters than training examples. 
    It is believed that implicit bias is a key factor in their ability to generalize, and hence it was widely studied in recent years. 
    In this short survey, we explain the notion of implicit bias, review main results and discuss their implications.
\end{abstract}

\section{Introduction}

\emph{Deep learning} has been highly successful in recent years and has led to dramatic improvements in multiple domains. 
Deep-learning algorithms often \emph{generalize} quite well in practice, namely, given access to 
labeled
training data, they return neural networks that correctly label unobserved test data. However, despite much research our theoretical understanding of generalization in deep learning is still limited.

Neural networks used in practice often have far more learnable parameters than training examples. In such \emph{overparameterized} settings, one might expect \emph{overfitting} to occur, that is, the learned network might perform well on the training dataset and perform poorly on test data. 
Indeed, in overparameterized settings, there are many solutions that perform well on the training data, but most of them do not generalize well. Surprisingly, it seems that gradient-based deep-learning algorithms\footnote{Neural networks are trained using gradient-based algorithms, where the network's parameters are randomly initialized, and then adjusted in many stages in order to fit the training dataset by using information 
based on the gradient of a loss function w.r.t. the network parameters.} 
prefer the solutions that generalize well~\citep{zhang2021understanding}.

\begin{figure}[t]
  \centering
  \includegraphics[scale=0.38]{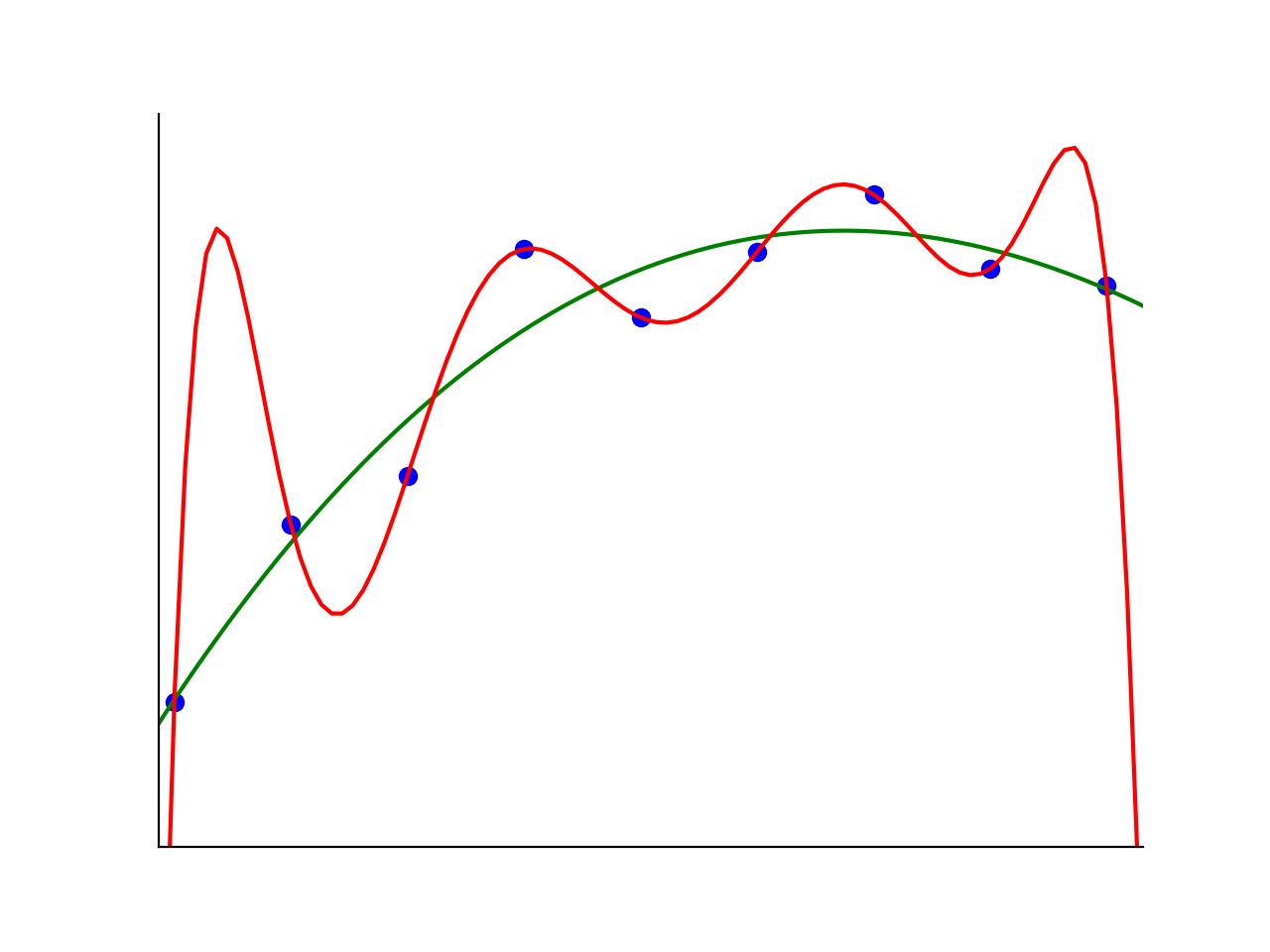}
  \caption{Fitting training data with a degree-$3$ polynomial (green) and degree-$8$ polynomial (red). The latter achieves better accuracy on the training data but it overfits.}
  \label{fig:overfit}
\end{figure}

Decades of research in learning theory suggest that in order to avoid overfitting one should use a model which is ``not more expressive than necessary''. Namely, the model should be able to 
perform well on the training data,
but should be as ``simple'' as possible. This idea goes back to the \emph{Occam's Razor} philosophical principle, which argues that we should prefer simple explanations over complicated ones. 
For example, in \figref{fig:overfit} the data points are labeled according to a degree-$3$ polynomial plus a small random noise, and we fit it with a degree-$3$ polynomial (green) and with a degree-$8$ polynomial (red). The degree-$8$ polynomial achieves better accuracy on the training data, but it overfits and will not generalize well. 

Simplicity in neural networks may stem from having a small number of parameters (which is often not the case in modern deep learning), but may also be achieved by adding a \emph{regularization term} during the training, which encourages networks that minimize a certain measure of complexity. For example, we may add a regularizer that penalizes models where the Euclidean norm of the parameters (viewed as a vector) is large. 
However, in practice neural networks seem to generalize well even when trained without such an explicit regularization \citep{zhang2021understanding}, and hence the success of deep learning cannot be attributed to explicit regularization.
Therefore,
it is believed that gradient-based algorithms induce an \emph{implicit bias} (or \emph{implicit regularization})~\citep{
neyshabur2017exploring} which prefers solutions that generalize well, and characterizing this bias has been a subject of extensive research.

In this review article, we discuss the implicit bias in training neural networks using gradient-based methods. The literature on this subject has rapidly expanded in recent years, and we aim to provide a high-level overview of some fundamental results. This article is not a comprehensive survey, and there are certainly important results which are not discussed here. 

\section{The double-descent phenomenon}

\begin{figure*}[t]
  \includegraphics[width=\textwidth]{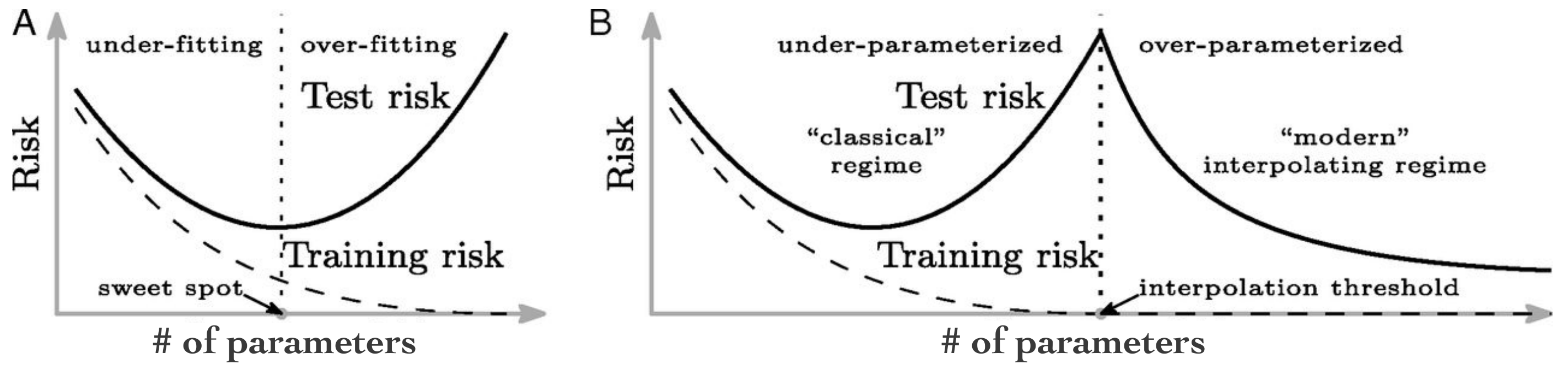}
  \caption{The double-descent phenomenon. Dashed lines denote the training risk and solid lines denote the test risk.  
  (A) The classical U-shaped curve. (B) The double-descent curve, which extends the U-shaped curve with a second descent. Beyond the ``interpolation threshold'' the model fits the training data perfectly, but the test risk decreases. 
  The figure is from \cite{belkin2019reconciling}.}
  \label{fig:double}
\end{figure*}

An important implication of the implicit bias in deep learning is the \emph{double-descent} phenomenon, observed by Belkin et al. \citep{belkin2019reconciling}. As we already discussed, conventional wisdom in machine learning suggests that the number of parameters in the neural network should not be too large in order to avoid overfitting 
(when training without explicit regularization). Also, it should not be too small, in which case the model is not expressive enough and hence it performs poorly even on the training data, a situation called \emph{underfitting}. 
This classical thinking can be captured by the U-shaped risk (i.e., error or loss) curve from Figure~\ref{fig:double}(A). Thus, as we increase the number of parameters, the training risk decreases, and the test risk initially decreases and then increases. Hence, there is a ``sweet spot'' where the test risk is minimized. This classical U-shaped curve suggests that we should not use a model that perfectly fits the training data.

However, it turns out that the U-shaped curve only provides a partial picture. If we continue to increase the number of parameters in the model after the training set is already perfectly labeled, then there might be a second descent in the test risk (hence the name ``double descent''). As can be seen in Figure~\ref{fig:double}(B), by increasing the number of parameters beyond what is required to perfectly fit the training set, the generalization improves. Hence, in contrast to the classical approach which seeks a sweet spot, we may achieve generalization by using overparameterized neural networks. Modern deep learning indeed relies on using highly overparameterized networks. 

The double-descent phenomenon is believed to be a consequence of the implicit bias in deep-learning algorithms. When using overparameterized models, gradient methods converge to networks that generalize well by implicitly minimizing a certain measure of the network's complexity. As we will see next, characterizing this complexity measure in different settings is a challenging puzzle.

\section{Implicit bias in classification}

We start with the implicit bias 
in classification tasks, namely, where the labels are in $\{-1,1\}$. 
Neural networks are trained in practice using the \emph{gradient-descent} algorithm and its variants, such as \emph{stochastic gradient descent} (SGD). In gradient descent, we start from some random initialization $\btheta_0$ of the network's parameters, and then in each iteration $t \geq 1$ we update $\btheta_t = \btheta_{t-1} - \eta \nabla \cl(\btheta_{t-1})$, where $\eta>0$ is the step size and $\cl$ is some \emph{empirical loss} that we wish to minimize. Here, we will focus on \emph{gradient flow}, which is a continuous version of gradient descent. That is, we train the parameters $\btheta$ of the considered neural network, such that for all time $t \geq 0$ we have $\frac{d \btheta(t)}{dt} = -\nabla \cl(\btheta(t))$, where $\cl$ is the empirical loss, and $\btheta(0)$ is some random initialization.\footnote{If $\cl$ is non-differentiable, e.g., in ReLU networks, then we use the \emph{Clarke subdifferential}, which is a generalization of the derivative for non-differentiable functions.} We focus on the \emph{logistic loss function} (a.k.a. \emph{binary cross entropy}):
For a training dataset $\{(\bx_i,y_i)\}_{i=1}^n \subseteq \reals^d \times \{-1,1\}$ and a neural network $\cn_\btheta:\reals^d \to \reals$ parameterized by $\btheta$, the empirical loss is defined by $\cl(\btheta) = \frac{1}{n}\sum_{i=1}^n \log\left(1 + e^{-y_i \cdot \cn_\btheta(\bx_i)} \right)$.
We note that gradient flow 
corresponds to gradient descent with an infinitesimally small step size,
and many implicit-bias results are shown for gradient flow since it is often easier to analyze.

\subsection{Logistic regression}

Logistic regression is 
perhaps the simplest classification setting
where implicit bias of neural networks 
can be considered. It corresponds to a neural network with a single neuron that does not have an activation function. That is, the model here is a linear predictor $\bx \mapsto \bw^\top \bx$, where $\bw \in \reals^d$ are the learned parameters. 

We assume that the data is \emph{linearly separable}, i.e., there exists some $\hat{\bw} \in \reals^d$ such that for all $i \in \{1,\ldots,n\}$ we have $y_i \cdot \hat{\bw}^\top \bx_i > 0$.
Note that $\cl(\bw)$ is strictly positive for all $\bw$, but it may tend to zero as $\norm{\bw}_2$ tends to infinity. For example, for $\alpha \in \reals$ and $\bw=\alpha \hat{\bw}$ (where $\hat{\bw}$ is the vector defined above) we have $\lim_{\alpha \to \infty} \cl(\bw) = 0$. However, there might be infinitely many directions $\hat{\bw}$ that satisfy this condition. Interestingly, Soudry et al. \citep{soudry2018implicit} showed that gradient flow converges in direction to the $\ell_2$ \emph{maximum-margin predictor}. That is, $\tilde{\bw}=\lim_{t \to \infty}\frac{\bw(t)}{\norm{\bw(t)}_2}$ exists, and $\tilde{\bw}$ points at the direction of the maximum-margin predictor, defined by 
\[
    \argmin_{\bw \in \reals^d} \norm{\bw}_2 \;\;\;\; \text{s.t. } \;\;\; \forall i \in \{1,\ldots,n\} \;\; y_i \cdot \bw^\top \bx_i \geq 1~.
\]

The above characterization of the implicit bias in logistic regression can be used to explain generalization. Consider the case where $d > n$, thus, the input dimension is larger than the size of the dataset, and hence there are infinitely many 
vectors $\bw$ with $\norm{\bw}_2=1$ such that for all $i \in \{1,\ldots,n\}$ we have $y_i \cdot \bw^\top \bx_i > 0$, i.e., there are infinitely many directions that correctly label the training data. 
Some of the directions which fit the training data generalize well, and others do not. The above result guarantees that gradient flow converges to the maximum-margin direction. Consequently, we can explain generalization by using standard margin-based generalization bounds (cf. \cite{shalev2014understanding}), which imply that predictors achieving large margin generalize well.

\subsection{Linear networks}

We turn to
deep 
neural networks where the activation is the identity function. Such neural networks are called \emph{linear networks}. 
These networks compute linear functions, but the network architectures induce significantly different dynamics of gradient flow compared to the case of linear predictors $\bx \mapsto \bw^\top \bx$ that we already discussed. 
Hence, the implicit bias in linear networks has been extensively studied, as a first step towards understanding implicit bias in deep nonlinear networks.  
It turns out that understanding implicit bias in linear networks is highly non-trivial, and its study reveals valuable insights.

A linear \emph{fully-connected} 
network $\cn:\reals^d \to \reals$ of depth $k$ computes a function $\cn(\bx) = W_k \ldots W_1 \bx$, where $W_1,\ldots,W_k$ are weight matrices (where $W_k$ is a row vector). 
The trainable parameters are a collection $\btheta = [W_1,\ldots,W_k]$ of the weight matrices. 
By \cite{ji2020directional},\footnote{Similar results under stronger assumptions were previously established in \cite{gunasekar2018bimplicit,ji2018gradient}.} if gradient flow converges to zero loss, namely, $\lim_{t \to \infty}\cl(\btheta(t)) = 0$, then the vector $W_1 \cdot \ldots \cdot W_k$ converges in direction to the $\ell_2$ maximum-margin predictor. 
That is, although the dynamics of gradient flow in the case of a deep fully-connected linear network is significantly different compared to the case of a linear predictor, 
in both cases gradient flow is biased towards the $\ell_2$ maximum-margin predictor. We also note that by \cite{ji2018gradient}, the weight matrices $W_1,\ldots,W_{k-1}$ converge to matrices of rank $1$. Thus, gradient flow on fully-connected linear networks is also biased towards rank minimization of the weight matrices.

Once we consider linear networks which are not fully connected, gradient flow no longer maximizes the $\ell_2$ margin. For example, Gunasekar et al. \citep{gunasekar2018bimplicit} showed that in \emph{linear diagonal networks} (i.e., where the weight matrices $W_1,\ldots,W_{k-1}$ are diagonal) gradient flow encourages predictors that maximize the margin w.r.t. the $\norm{\cdot}_{2/k}$ quasi-norm. As a result, diagonal networks are biased towards sparse linear predictors. In \emph{linear convolutional networks} they proved bias towards sparsity of the linear predictors in the frequency domain (see also \cite{yun2020unifying}).  

\subsection{Homogeneous neural networks}

Neural networks of practical interest have nonlinear activations and compute nonlinear predictors. 
The notion of margin maximization in nonlinear predictors is generally not well-defined. 
Nevertheless, it has been established that for certain neural networks gradient flow maximizes the margin in \emph{parameter space}. 
Thus, while in linear networks we considered margin maximization in \emph{predictor space} (a.k.a. \emph{function space}), here we consider margin maximization w.r.t. the network's parameters.

Consider a neural network with parameters $\btheta$, denoted by $\cn_\btheta:\reals^d \to \reals$. We say that $\cn_\btheta$ is \emph{homogeneous} if there exists $L>0$ such that for every $\alpha>0$ and $\btheta,\bx$ we have $\cn_{\alpha \cdot \btheta}(\bx) = \alpha^L \cn_\btheta(\bx)$. 
Here we think about $\btheta$ as a vector obtained by concatenating all the parameters.
That is, in homogeneous networks, scaling the parameters by any factor $\alpha>0$ scales the predictions by $\alpha^L$. We note that a feedforward neural network with the ReLU activation (namely, $z \mapsto \max\{0,z\}$) is homogeneous if it does not contain skip-connections (i.e., residual connections), and does not have bias terms, except maybe for the first hidden layer.
Also, we note that a homogeneous network might include convolutional layers.
In papers by Lyu and Li \citep{lyu2019gradient} and by Ji and Telgarsky \citep{ji2020directional},\footnote{A similar result under stronger assumptions was previously established in \cite{nacson2019lexicographic}.} it was shown that if gradient flow on homogeneous networks reaches a sufficiently small loss (at some time $t_0$), then as $t \to \infty$ it converges to zero loss, the parameters converge in direction, i.e., $\tilde{\btheta}=\lim_{t \to \infty}\frac{\btheta(t)}{\norm{\btheta(t)}_2}$ exists, and $\tilde{\btheta}$ is biased towards margin maximization in the following sense. Consider the following margin maximization problem in parameter space:
\[
    \min_\btheta \frac{1}{2} \norm{\btheta}_2^2 \;\;\;\; \text{s.t. } \;\;\; \forall i \in \{1,\ldots,n\} \;\; y_i \cn_\btheta(\bx_i) \geq 1~.
\]
Then, $\tilde{\btheta}$ points at the direction of a first-order stationary point of the above optimization problem, which is also called \emph{Karush–Kuhn–Tucker point}, or \emph{KKT point} for short. 
The KKT approach allows inequality constraints and is a generalization of the method of \emph{Lagrange multipliers}, which allows only equality constraints.

A KKT point satisfies several conditions (called \emph{KKT conditions}), and it was proved in \cite{lyu2019gradient} that in the case of homogeneous neural networks these conditions are necessary for optimality. However, they are not sufficient even for local optimality (cf. \cite{vardi2021margin}). Thus, a KKT point may not be an actual optimum of the maximum-margin problem.
It is analogous to showing that some unconstrained optimization problem converges to a point where the gradient is zero, without proving that it is a global/local minimum.
Intuitively, convergence to a KKT point of the maximum-margin problem implies a certain bias towards margin maximization in parameter space, although it does not guarantee convergence to a maximum-margin solution. 

As we already discussed, in linear classifiers and linear neural networks, margin maximization (in predictor space) can explain generalization using well-known margin-based generalization bounds. Can margin maximization in parameter space 
explain generalization in nonlinear neural networks? In recent years several works showed margin-based generalization bounds for neural networks (e.g., \cite{neyshabur2015norm,bartlett2017spectrally,golowich2018size,vardi2022sample}). Hence, generalization in neural networks might be established by combining these results with results on the implicit bias towards margin maximization in parameter space. On the flip side, we note that it is still unclear how tight the known margin-based generalization bounds for neural networks are, and whether the sample complexity implied by such results (i.e., the required size of the training dataset) may be small enough to capture the situation in practice. 

Several results on implicit bias were shown for some specific cases of nonlinear homogeneous networks. For example: 
\cite{chizat2020implicit} 
showed bias towards margin maximization w.r.t. a certain function norm (known as the variation norm) in infinite-width two-layer homogeneous networks;
\cite{lyu2021gradient} proved margin maximization in two-layer
Leaky-ReLU networks trained on linearly separable and symmetric
data, and \cite{sarussi2021towards,frei2022implicit} proved convergence to a linear classifier in two-layer Leaky-ReLU networks under different assumptions; 
\cite{safran2022effective} showed bias towards minimizing the number of linear regions in univariate two-layer ReLU networks (see also \cite{ergen2021convex,savarese2019infinite}).

Finally, the implicit bias in \emph{non-homogeneous neural networks} (such as ResNets) is currently not well-understood.\footnote{We remark that a result by \cite{nacson2019lexicographic} suggests that for a sum of homogeneous networks of different orders (such a sum is non-homogeneous), the implicit bias may encourage solutions that discard the networks with the smallest order.}
Improving our knowledge of non-homogeneous networks is an important challenge in the path towards understanding implicit bias in deep learning.


\subsection{Extensions}

For simplicity, we considered so far only gradient flow in binary classification. We note that many of the above results can also be extended to other gradient methods (such as gradient descent, steepest descent and SGD), and to multiclass classification with the cross-entropy loss (see, e.g., \cite{soudry2018implicit,lyu2019gradient,gunasekar2018characterizing,nacson2019stochastic}).

The margin-maximization guarantees that we discussed for gradient flow hold in an asymptotic sense, and 
an important question is how fast the convergence is. It turns out that the convergence rate might be extremely slow \citep{soudry2018implicit,nacson2019convergence,ji2018risk,ji2021characterizing,nacson2019stochastic,ji2021fast}. 
For example, when learning a linear predictor on linearly-separable training dataset using gradient descent (with a sufficiently small step size), after $T$ iterations the distance between the normalized predictor $\bar{\bw}_T$ and the maximum margin predictor $\bw^*$ generally satisfies $\norm{\bar{\bw}_T - \bw^*}_2 = \co\left(1/\log(T)\right)$,\footnote{See \cite{soudry2018implicit} for a more precise statement.} and hence to reach $\norm{\bar{\bw}_T - \bw^*}_2 \leq \gamma$ for some $\gamma>0$, the number of iterations must be exponential in $1/\gamma$.
We remark that Shamir \citep{shamir2021gradient} showed that the slow convergence rate to the maximum-margin predictor does not imply that 
$T$ must be extremely large in order to avoid overfitting.
Namely, he proved that also for much smaller values of $T$, the predictor $\bar{\bw}_T$ achieves a sufficiently large margin on a sufficiently large portion of the dataset, which implies good generalization properties. 

Moreover, so far we considered only training with the logistic loss, which is a common loss function for binary classification. The results can generally 
be extended to 
loss functions that have an exponential tail, but the implicit bias is different when using other loss functions (e.g., losses with a polynomial tail)~\cite{ji2020gradient,ji2021characterizing,nacson2019convergence}.

Finally, all the results described so far do not depend on the initialization of gradient flow. For example, when training homogeneous networks, as the time $t$ tends to infinity, gradient flow converges to a KKT point of the maximum-margin problem (assuming that the loss 
reaches a sufficiently small value), regardless of the initialization.
Namely, for every initialization where gradient flow reaches small loss, it is guaranteed to converge to a KKT point.
However, such an asymptotic analysis only provides a partial picture. The authors in \cite{moroshko2020implicit} considered certain
linear diagonal networks
of depth $k$, and showed that if we train the network only until we reach some fixed accuracy, e.g., until the loss reaches $10^{-10}$ (instead of considering $t \to \infty$), then the initialization plays a crucial role: 
If the initialization scale is large w.r.t. the considered accuracy (a setting which corresponds to the so-called \emph{kernel regime}),
then the implicit bias is given by the $\ell_2$ norm, rather than the $\ell_{2/k}$ quasi-norm in the 
setting
studied so far. 
Thus, 
to understand implicit bias in practice, the asymptotic results may not suffice, and we may need to take the initialization of gradient flow into account.

Additional aspects of the implicit bias, which apply both to classification and regression, are discussed in Sections~\ref{sec:dynamical} and~\ref{sec:additional}.

\section{Implicit bias in regression}

We turn to consider the implicit bias of gradient methods in regression tasks, namely, where the labels are in $\reals$. 
We focus on gradient flow and on the \emph{square-loss} function. Thus, we have $\frac{d \btheta(t)}{dt} = -\nabla \cl(\btheta(t))$, where the empirical loss $\cl(\btheta)$ is defined such that given
a training dataset $\{(\bx_i,y_i)\}_{i=1}^n \subseteq \reals^d \times \reals$ and a neural network $\cn_\btheta:\reals^d \to \reals$, we have $\cl(\btheta) = \frac{1}{n} \sum_{i=1}^n \left( \cn_\btheta(\bx_i) - y_i \right)^2$. In an overparameterized setting, there might be many possible choices of $\btheta$ such that $\cl(\btheta)=0$, thus there are many global minima. Ideally, we want to find some implicit regularization function $\calr(\btheta)$ such that gradient flow prefers global minima which minimize $\calr$. That is, if gradient flow converges to some $\btheta^*$ with $\cl(\btheta^*)=0$, then we have $\btheta^* \in \argmin_\btheta \calr(\btheta)$ s.t. $\cl(\btheta)=0$.

\subsection{Linear regression}

We start with linear regression, which is perhaps the simplest regression setting where the implicit bias of neural networks can be considered. Here, the model is a linear predictor $\bx \mapsto \bw^\top \bx$, where $\bw \in \reals^d$ are the learned parameters. In this case, it is not hard to show that gradient flow converges to the global minimum of $\cl(\bw)$, which is closest to the initialization $\bw(0)$ in $\ell_2$ distance \citep{zhang2021understanding,gunasekar2018characterizing}. Thus, the implicit regularization function is $\calr_{\bw(0)}(\bw) = \norm{\bw - \bw(0)}_2$. Minimizing this $\ell_2$ norm allows us to explain generalization using standard norm-based generalization bounds \citep{shalev2014understanding}. 

We note that the results on linear regression can be extended to optimization methods other than gradient flow, such as gradient descent, SGD, and \emph{mirror descent} \cite{gunasekar2018characterizing,azizan2018stochastic}.\footnote{For mirror descent with a potential function $\psi(\bw)$, the implicit bias is given by $\calr(\bw) = D_\psi\left(\bw,\bw(0)\right)$, where $D_\psi$ is the \emph{Bregman divergence} w.r.t. $\psi$. In particular, if we start at $\bw(0)=\argmin_\bw \psi(\bw)$ then we have $\calr(\bw) = \psi(w)$.}

\subsection{Linear networks}

A linear neural network $\cn_\btheta:\reals^d \to \reals$ has the identity activation function and computes a linear predictor $\bx \mapsto \bw^\top \bx$, where $\bw$ is the vector obtained by multiplying the weight matrices in $\cn_\btheta$. Hence, instead of seeking an implicit regularization function $\calr(\btheta)$ we may aim to find $\calr(\bw)$.
Similarly to our discussion in the context of classification, 
the network architectures in deep linear networks induce significantly different dynamics of gradient flow compared to the case of linear regression. 
Hence, 
the implicit bias in such networks has been studied as a first step towards understanding implicit bias in deep nonlinear networks.

Exact expressions for $\calr(\bw)$ have been obtained for linear diagonal networks and linear 
fully-connected networks \citep{azulay2021implicit,yun2020unifying,woodworth2020kernel}. The expressions for $\calr(\bw)$ are rather complicated, and depend on the initialization scale, i.e., the norm of $\bw(0)$, and the initialization ``shape'', namely, the relative magnitudes of different weights and layers in the network. Below we discuss a few special cases.

Recall that diagonal networks are neural networks where the weight matrices are diagonal. The regularization function $\calr$ has been obtained for a few variants of diagonal networks. In \cite{woodworth2020kernel} the authors considered networks of the form $\bx \mapsto (\bu_+ \circ \bu_+ - \bu_- \circ \bu_-)^\top \bx$, where $\bu_+, \bu_- \in \reals^d$ and the operator $\circ$ denotes the Hadamard (entrywise) product. Thus, in this network, which can be viewed as a variant of a diagonal network, the weights in the two layers are shared. For an initialization $\bu_+(0) = \bu_-(0) = \alpha \cdot \one$, the scale $\alpha>0$ of the initialization does not affect $\bw(0) = \bu_+(0) \circ \bu_+(0) - \bu_-(0) \circ \bu_-(0) = \zero$. However, the initialization scale does affect the implicit bias: if $\alpha \to 0$ then $\calr(\bw) = \norm{\bw}_1$ and if $\alpha \to \infty$ (which corresponds to the so-called \emph{kernel regime}) then $\calr(\bw) = \norm{\bw}_2$. Thus, the implicit bias transitions between the $\ell_1$ norm and the $\ell_2$ norm as we increase the initialization scale. A similar result holds also for deeper networks. 
In two-layer ``diagonal networks'' with a similar structure, but where the weights are not shared, \cite{azulay2021implicit} showed that the implicit bias is affected by both the initialization scale and ``shape''. 
In \cite{yun2020unifying} the authors showed  
a transition between the $\ell_1$ and $\ell_2$ norms
both in diagonal networks and in certain convolutional networks (in the frequency domain). 

In two-layer fully-connected linear networks with infinitesimally small initialization, gradient flow minimizes the $\ell_2$ norm, namely, $\calr(\bw)=\norm{\bw}_2$ \citep{azulay2021implicit,yun2020unifying}. 
This result was shown also for deep fully-connected linear networks, under certain assumptions on the initialization \citep{yun2020unifying}.
These results were extended to a \emph{fine-tuning} setup in \cite{shachaf2021theoretical}.

Some additional aspects of the implicit bias in linear networks are discussed in Sections~\ref{sec:dynamical} and~\ref{sec:additional}.

\subsection{Matrix factorization as a test-bed for neural networks}

Towards the goal of understanding implicit bias in neural networks, much effort was directed at the \emph{matrix factorization} (and more specifically, \emph{matrix sensing}) problem. In this problem, we are given a set of observations about an unknown matrix $W^* \in \reals^{d \times d}$ and we find a matrix $W = W_2 W_1$ with $W_1,W_2 \in \reals^{d \times d}$ that is compatible with the given observations by running gradient flow. More precisely, consider the observations $\{(X_i,y_i)\}_{i=1}^n \subseteq \reals^{d \times d} \times \reals$ such that $y_i = \inner{W^*,X_i}$, where in the inner product we view $W^*$ and $X_i$ as vectors (namely, $\inner{W,X} = \text{tr}(W^\top X)$). Then, we find $W = W_2 W_1$ using gradient flow on the objective $\cl(W_1,W_2) = \frac{1}{n} \sum_{i=1}^n \left( \inner{W_2 W_1, X_i} - y_i \right)^2$. Since $W_1,W_2 \in \reals^{d \times d}$ then the rank of $W$ is not limited by the parameterization. However, the parameterization has a crucial effect on the implicit bias. Assuming that there are many matrices $W$ that are compatible with the observations, the implicit bias controls which matrix gradient flow finds. 
A notable special case of matrix sensing is \emph{matrix completion}. Here, every observation has the form $X_i = \be_{p_i} \be_{q_i}^\top$, where $\{\be_1,\ldots,\be_d\}$ is the standard basis. Thus, each observation $(X_i,y_i)$ reveals a single entry from the matrix $W^*$.

The matrix factorization problem is analogous 
to training a two-layer linear network. Furthermore, we may consider \emph{deep matrix factorization} where we have $W = W_k W_{k-1} \ldots W_1$, which is analogous to training a deep linear network. Therefore, this problem is considered a natural test-bed to investigate implicit bias in neural networks, and has been studied extensively in recent years (e.g., \cite{gunasekar2017implicit,li2018algorithmic,arora2019implicit,razin2020implicit,belabbas2020implicit,li2020towards}).

Gunasekar et al. \citep{gunasekar2017implicit} conjectured that in matrix factorization, the implicit bias of gradient flow starting from small initialization is given by the nuclear norm of $W$ (a.k.a. the trace norm). That is, $\calr(W)=\norm{W}_*$. They also proved this conjecture in the restricted case where the matrices $\{X_i\}_{i=1}^n$ commute.
The conjecture was further studied in
a line of works (e.g., \cite{li2018algorithmic,belabbas2020implicit,arora2019implicit,razin2020implicit}) providing positive and negative evidence, and was formally refuted by \cite{li2020towards}.
In \cite{razin2020implicit} the authors showed that gradient flow in matrix completion might approach a global minimum at infinity rather than converging to a global minimum with a finite norm. This result suggests that the implicit bias in matrix factorization may not be expressible by any norm or quasi-norm. 
They conjectured that the implicit bias can be explained by rank
minimization rather than norm minimization. 
In \cite{li2020towards}, the authors
provided theoretical and empirical evidence that gradient flow with infinitesimally small initialization in the matrix-factorization problem is mathematically equivalent to a simple heuristic rank-minimization algorithm called \emph{Greedy Low-Rank Learning} (see also \cite{jacot2021deep}). 
This result was generalized to tensor factorization in \cite{razin2021implicit,razin2022implicit}.

Overall, although an explicit expression for a function $\calr$ is not known in the case of matrix factorization, we can view the implicit bias as a heuristic for rank minimization. 
It is still unclear what the implications of these results are for more practical nonlinear neural networks.

\subsection{Nonlinear networks}

Once we consider nonlinear networks the situation is even less clear. 
For a single-neuron network, namely, $\bx \mapsto \sigma(\bw^\top \bx)$, where $\sigma:\reals \to \reals$ is strictly monotonic, gradient flow converges to the closest global minimum in $\ell_2$ distance, similarly to the case of linear regression. Thus, we have $\calr_{\bw(0)}(\bw) = \norm{\bw - \bw(0)}_2$. 
However, if $\sigma$ is the ReLU activation then the implicit bias is not expressible by any non-trivial function of $\bw$. A bit more precisely, suppose that $\calr:\reals^d \to \reals$ is such that if gradient flow starting from $\bw(0)=\zero$ converges to a zero-loss solution $\bw^*$, then $\bw^*$ is a zero-loss solution that minimizes $\calr$. Then, such a function $\calr$ must be constant in $\reals^d \setminus \{\zero\}$. Hence, the approach of precisely specifying the implicit bias of gradient flow via a regularization function is not feasible in single-neuron ReLU networks \citep{vardi2021implicit}. It suggests that this approach may not be useful for understanding implicit bias in the general case of ReLU networks.

On the positive side, in single-neuron ReLU networks the implicit bias can be expressed approximately, within a factor of
$2$, by the $\ell_2$ norm. Namely, let $\bw^*$ be a zero-loss solution with a minimal $\ell_2$ norm, and assume that gradient flow converges to some zero-loss solution $\bw(\infty)$, then $\norm{\bw(\infty)}_2 \leq 2 \norm{\bw^*}_2$ \citep{vardi2021implicit}. 
  
For two-layer Leaky-ReLU networks with a single hidden neuron, an explicit expression for $\calr$ was obtained in \cite{azulay2021implicit}. However, if we replace the Leaky-ReLU activation with ReLU, then the implicit bias is not expressible by any non-trivial function, similarly to the case of a single-neuron ReLU network \citep{vardi2021implicit}.
A possible approach to circumvent this negative result it to make some assumptions on the training dataset. In \cite{boursier2022gradient}, the authors showed that if the training examples are orthogonal, then gradient flow on two-layer ReLU networks, under certain assumptions on the initialization, converges to a zero-loss solution that minimizes the $\ell_2$ norm of the parameters.

The results on implicit bias in matrix factorization might suggest that there is a certain tendency towards rank minimization in deep learning with the square loss. However, the authors in \cite{timor2022implicit} showed that gradient flow is not biased towards rank minimization of the weight matrices in ReLU networks, at least in the simple case of two-layer networks and small datasets.\footnote{We remark that \cite{timor2022implicit} showed a certain tendency towards rank minimization in deep networks trained with the square loss using weight decay, or with the logistic loss.}

Overall, our understanding of implicit bias in nonlinear networks with regression losses is very limited. While in classification we have a useful characterization of the implicit bias in nonlinear homogeneous models, here we do not understand even extremely simple models. Improving our understanding of this question is a major challenge for the upcoming years. 

In what follows, we will discuss some additional aspects of the implicit bias, which apply both to regression and classification.

\section{Dynamical stability} \label{sec:dynamical}

In the previous sections, we mostly focused on implicit bias in gradient flow, and also discussed cases where the results on gradient flow can be extended to other gradient methods.
However, when considering gradient descent or SGD with a finite step size (rather than infinitesimal), the discrete nature of the algorithms and the stochasticity induce additional implicit biases that do not exist in gradient flow. These biases are crucial for understanding the behavior of gradient methods in practice, as empirical results suggest that increasing the step size and decreasing the batch size may improve generalization (cf. \cite{keskar2016large,jastrzkebski2017three}).

It is well known that gradient descent and SGD cannot stably converge to minima that are too sharp relative to the step size (see, e.g., \cite{mulayoff2021implicit,nacson2022implicit,mulayoff2020unique,wu2018sgd}). Namely, in a stable minimum, the maximal eigenvalue of the Hessian is at most $2/\eta$, where $\eta$ is the step size. As a result, when using an appropriate step size we can rule out stable convergence to certain (sharp) minima, and thus encourage convergence to flat minima, which are believed to generalize better \citep{keskar2016large,wu2018sgd}. Similarly, small batch sizes in SGD also encourage flat minima~\citep{keskar2016large,wu2018sgd,ma2021linear,wu2022does}. 

We note that a function might be represented with many different networks, i.e., there are many sets of parameters that correspond to the same function. However, it is possible that some of these representations are sharp (in parameter space) and others are flat (cf. \cite{dinh2017sharp}). As a result, understanding the implications in function space of the bias towards flat minima (in parameter space) is often a challenging question.

Implications of the implicit bias towards flat minima were studied in several works \citep{mulayoff2021implicit,nacson2022implicit,ma2021linear,wang2021large,wu2018sgd,mulayoff2020unique,nar2018step}.
For example, 
\cite{mulayoff2021implicit} studied flat minima in univariate ReLU networks, and showed that SGD with large step size is biased towards functions whose second derivative (w.r.t. the input) has a bounded weighted $\ell_1$ norm, which encourages convergence to smooth functions;
\cite{nacson2022implicit} showed that gradient descent with large step size on a diagonal linear network with non-centered data minimizes the $\ell_1$ norm of the corresponding linear predictor, even in settings where gradient descent with small step size minimizes the $\ell_2$ norm;
\cite{ma2021linear} studied minima stability in SGD, and showed that flat minima induce regularization of the function's gradient (w.r.t. the input data), and more generally that SGD regularizes Sobolev seminorms of the function; \cite{mulayoff2020unique,wang2021large} studied flat minima in linear neural networks and in matrix factorization, and showed bias towards solutions where the magnitudes of the layers are balanced, as well as other properties of flat solutions.
We note that all the above papers consider regression with the square loss.

An intriguing related phenomenon, observed by Cohen et al. \citep{cohen2021gradient}, is the tendency of gradient descent to operate in a regime called \emph{the Edge of Stability (EoS)}. They showed empirically for several architectures and tasks, and for both the cross-entropy and the square loss, that the behavior of gradient descent with step size $\eta$ consists of two stages. First, the loss curvature grows until the sharpness touches the bound of $2/\eta$. Then, gradient decent enters the EoS regime, where curvature hovers around this bound, and the train loss behaves non-monotonically, yet consistently decreases over long timescales. This phenomenon is not well understood, as traditional convergence analyses of gradient descent do not apply when the sharpness is above $2/\eta$. 
The study of the EoS regime may contribute to our understanding of the dynamics and implicit bias in gradient descent and SGD. 
It was investigated in several recent theoretical works \citep{arora2022understanding,chen2022gradient,ahn2022understanding,ma2022beyond,lyu2022understanding}. 
For example, \cite{arora2022understanding} showed that training with modified gradient descent or modified loss (for a large family of losses) provably enters the EoS regime, and that the loss decreases in a non-monotone manner, and \cite{chen2022gradient} studied EoS in training a two-layer single-neuron ReLU network with the square loss.

\section{Additional implicit biases} \label{sec:additional}



Additional aspects of the implicit bias have also been studied in the literature. Below we briefly discuss some of them. 

First, 
the implicit bias of SGD is 
affected when training with label noise, namely, when the training labels are perturbed by independent noise at each iteration \citep{blanc2020implicit,haochen2021shape,damian2021label,li2021happens,pillaud2022label}.
For example,
in \cite{haochen2021shape,pillaud2022label} it is shown that when training a linear diagonal network using SGD with label noise, there is bias towards sparse solutions even when the initialization scale is large. This is in contrast to 
noise-free training
where such initializations 
lead to implicit $\ell_2$ regularization.

Another direction for understanding implicit bias in gradient descent and SGD, is to transform it into gradient flow on a modified loss.
In \cite{barrett2020implicit,smith2021origin}, the authors show that the discrete iterates of gradient descent and SGD with small step size lie close to the path of gradient flow on certain modified losses, obtained by adding explicit regularizers (see also \cite{pesme2021implicit}). 

An additional interesting aspect of the implicit bias in certain neural networks 
is a tendency towards balanced layers. 
In \cite{du2018algorithmic}, the authors proved that when training fully-connected neural networks with homogeneous activation functions using gradient flow with any differentiable loss function, we have $\frac{d}{dt}\left(\norm{W_j}_F^2 - \norm{W_{j+1}}_F^2 \right) = 0$ for all $j$, where $W_{j}$ and $W_{j+1}$ are the weight matrices in layers $j$ and $j+1$. Thus, the difference between the squared Frobenius norms of the layers remain invariant throughout training. Moreover, they showed that a similar result holds also for networks with sparse connections and shared weights, such as convolutional neural networks.
Note that when starting gradient flow from a small initialization, it implies that the Frobenius norms of the layers remain roughly balanced throughout training.
For linear neural networks an even stronger notion of balancedness holds: \cite{arora2018optimization,du2018algorithmic} proved that $\frac{d}{dt}\left(W_j W_j^\top- W_{j+1}^\top W_{j+1}\right) = 0$ for all $j$.
Furthermore, as we already mentioned, \cite{mulayoff2020unique,wang2021large} showed that in linear neural networks and matrix factorization, when training using gradient descent with large step size, the implicit bias towards flat solutions implies bias towards balanced layers.   
 
Many of the existing results on the implicit bias characterize properties of the parameters of the learned neural networks, namely, they consider the implicit bias in parameter space. However, since a function typically has many different representations, then understanding the implications of these results on function space is a non-trivial question. Several works studied the implications of norm minimization in parameter space on function space for different architectures of neural networks: \cite{savarese2019infinite,ergen2021convex} studied univariate two-layer neural networks and \cite{ongie2019function} extended the analysis to the multivariate case, \cite{gunasekar2018bimplicit} studied fully-connected, diagonal and convolutional linear networks, and \cite{jagadeesan2022inductive,pilanci2020neural,dai2021representation} studied larger families of convolutional linear networks.  

Finally, we note that in this article we mostly considered implicit bias in the \emph{rich regime}, rather than in the \emph{NTK regime} \citep{jacot2018neural} (a.k.a. \emph{kernel regime}). The NTK regime does not capture feature learning, which seems to be a crucial element in the 
success of deep learning. 
The implicit bias in the NTK regime 
has been 
studied in several works 
(see, e.g., \cite{williams2019gradient,bietti2019inductive}), 
but we leave the discussion on these results outside the scope of this~review.

\section{Implications beyond generalization}

While the primary motivation for studying implicit bias is to better understand generalization in deep learning, it may also have other significant implications. Indeed, various phenomena may stem from the tendency of gradient-based algorithms to prefer specific solutions over others. We believe that exploring such implications is an exciting research direction, and demonstrate it below with two examples.

First, neural networks are known to be extremely vulnerable to \emph{adversarial examples} \cite{szegedy2013intriguing}, namely, small perturbations to the inputs might change the network's predictions. This phenomenon has been widely studied, but it is still 
not well-understood.
Specifically, it is 
unclear
why gradient methods tend to learn \emph{non-robust networks}, namely, networks that are susceptible to adversarial examples, even in cases where robust networks exist. The tendency of gradient methods to learn non-robust networks can be viewed as an implication of the implicit bias. See \cite{vardi2022gradient} for some results in this direction.  

Second, the implicit bias might shed light on the hidden representations learned by neural networks, 
and on the extent to which neural networks are susceptible to privacy attacks. In \cite{haim2022reconstructing}, the authors used the characterization of the implicit bias in homogeneous networks due to \cite{lyu2019gradient,ji2020directional}, and showed that training data can be reconstructed from trained networks. Thus, neural networks ``memorize'' training data, and by using known properties of the implicit bias the data may be reconstructed, which might have negative implications on privacy.

\vspace{-0.08cm}

\section{Conclusion}

Deep-learning algorithms exhibit remarkable performance,
but it is not well-understood why they are able to generalize despite having much more parameters than training examples. Exploring generalization in deep learning is an intriguing puzzle with both theoretical and practical implications. It is believed that implicit bias is a main ingredient in the ability of deep-learning algorithms to generalize, and hence it has been widely studied.
Moreover, investigating the implicit bias might have important implications beyond~generalization.

Our understanding of implicit bias improved dramatically in recent years, but is still far from satisfactory. 
We believe that further progress in the study of implicit bias will eventually shed light on 
the mystery of generalization in deep learning.

\vspace{-0.07cm}

\subsection*{Acknowledgements}

I thank Nadav Cohen, Noam Razin, Ohad Shamir and Daniel Soudry for valuable comments and discussions.

\vspace{-0.07cm}

\bibliographystyle{abbrvnat}
\bibliography{bib}

\end{document}